\pdfoutput=1

\documentclass[11pt]{article}

\usepackage{float}
\usepackage[final]{acl}

\usepackage{times}
\usepackage{latexsym}

\usepackage[T1]{fontenc}

\usepackage[utf8]{inputenc}

\usepackage{microtype}

\usepackage{inconsolata}
\usepackage{multirow}
\usepackage{amsmath}
\usepackage{amssymb}
\usepackage{amsfonts}
\usepackage{multirow}
\usepackage{multicol}
\usepackage{array}
\usepackage{bm}
\usepackage{booktabs}
\usepackage{amsmath}
\usepackage{graphicx}
\usepackage{subfigure}
\usepackage[ruled]{algorithm2e}
\usepackage{enumitem}

\title{CogAtom: From Cognitive Atoms to Olympiad-level Mathematical Reasoning in Large Language Models}

\author{Zhuofan Chen\textsuperscript{1}\quad
        Jiyuan He\textsuperscript{2}\quad
        Yichi Zhang\textsuperscript{1}\quad
        Xing Hu\textsuperscript{2}\quad \\
        {\bf Haoxing Wen\textsuperscript{2}}\quad
        {\bf Jun Bai\textsuperscript{3}}\quad
        {\bf Wenge Rong\textsuperscript{1}}\\
  \textsuperscript{1} School of Computer Science and Engineering, Beihang University, China\\
  \textsuperscript{2} Meituan Inc., China\\
  \textsuperscript{3} Beijing Institute for General Artificial Intelligence, China\\
  \{zhuofanchen, yichizhang, w.rong\}@buaa.edu.cn,\\
  \{hejiyuan, huxing11, wenhaoxing\}@meituan.com, baijun@bigai.ai\\
}

\begin{document}
\maketitle
\begin{abstract}
Mathematical reasoning poses significant challenges for Large Language Models (LLMs) due to its demand for multi-step reasoning and abstract conceptual integration. While recent test-time scaling techniques rely heavily on high-quality, challenging problems, the scarcity of Olympiad-level math problems remains a bottleneck. We introduce CogAtom, a novel cognitive atom-based framework for synthesizing mathematically rigorous and cognitively diverse problems. Unlike prior approaches, CogAtom models problem construction as a process of selecting and recombining fundamental reasoning units, cognitive atoms, extracted from human-authored solutions. A diversity-promoting random walk algorithm enables exploration of the cognitive atom space, while a constraint-based recombination mechanism ensures logical soundness and structural validity. The combinatorial nature of the graph structure provides a near-infinite space of reasoning paths, and the walk algorithm systematically explores this space to achieve large-scale synthesis of high-quality problems; meanwhile, by controlling the number of cognitive atoms, we can precisely adjust problem difficulty, ensuring diversity, scalability, and controllability of the generated problems. Experimental results demonstrate that CogAtom outperforms existing methods in accuracy, reasoning depth, and diversity, generating problems that closely match the difficulty of AIME while exceeding it in structural variation. Our work offers a cognitively grounded pathway toward scalable, high-quality math problem generation.\footnote{Our code is publicly available at \url{https://github.com/Icarus-1111/CogAtom}.}
\end{abstract}

\section{Introduction}
\label{sec:introduction}
Reasoning abilities are core cognitive mechanisms underlying human problem-solving \citep{DBLP:journals/tamm/Hersh15}.
Among them, mathematical reasoning stands out for its unique cognitive complexity, requiring abstract concept comprehension, logical inference across domains, and multi-step solution strategies \citep{DBLP:conf/nips/HendrycksBKABTS21, DBLP:conf/nips/WangLXL24, DBLP:conf/iclr/YueQZFH00C24, DBLP:conf/nips/WeiSW24}. As a result, mathematical reasoning has become a key benchmark for evaluating the progress of Large Language Models (LLMs) toward Artificial General Intelligence (AGI) \citep{DBLP:conf/naacl/ZhongCGLLWSCD24}.

With the paradigm of LLMs shifting from training-time compute scaling toward test-time compute scaling \citep{DBLP:journals/corr/abs-2408-03314, DBLP:journals/corr/abs-2501-12948, DBLP:journals/corr/abs-2505-09388}, it heavily depends on the high-quality, multi-step mathematical problems as these high-quality problems are essential for effectively exposing model deficiencies in multi-step reasoning and validating the performance boundaries of test-time optimization strategies \citep{DBLP:conf/acl/LuZRWSPZL24, DBLP:conf/iclr/XuJNDP0L25}. This shift has generated a substantial demand for data that is simultaneously: (1) large-scale and scalable, to support compute-intensive optimization strategies; (2) highly challenging, to drive models beyond their current capabilities; and (3) cognitively diverse, to mitigate overfitting to familiar solution patterns. However, the existing collection of human-authored, Olympiad-level problems is static, limited in size, and does not fulfill these requirements. This pronounced scarcity of suitable data has become a primary obstacle to the rigorous development and evaluation of advanced reasoning systems \citep{DBLP:conf/nips/HendrycksBKABTS21, DBLP:conf/iclr/YueQZFH00C24, numinamath2024}.

The most straightforward solution to this data scarcity problem is to synthesize challenging problems automatically.
Existing synthesis methods can be categorized as prompt engineering-based methods \citep{DBLP:journals/corr/abs-2410-01560, DBLP:conf/iclr/YuJSYLZKLWL24, DBLP:conf/aaai/LiuZLY25}, corpus mining methods \citep{DBLP:conf/icml/ZhaoLK24, DBLP:conf/nips/YueZZC24}, evolutionary and transfer-based methods \citep{DBLP:conf/iclr/YuJSYLZKLWL24, DBLP:conf/iclr/LuoSX0LTGLCT025} and knowledge-driven synthesis methods \citep{DBLP:conf/icml/TangZWW24, DBLP:conf/aaai/HuangLGGSDC25, DBLP:journals/corr/abs-2503-02324}. However, these methods fail to model fundamental units of thought—cognitive atoms—and their combinatorial principles from a cognitive science perspective. In contrast, when mathematics experts design Olympiad problems, they carefully craft cognitive associations between concepts and construct rigorous logical frameworks that demand multi-level reasoning. Current methods are unable to effectively emulate this process, leading to substantial gaps between generated problems and human-authored Olympiad questions in terms of accuracy, logical coherence, and cognitive diversity.

To overcome these structural limitations, we introduce CogAtom, a framework that implements a paradigm shift from linear generation to structured synthesis, centered on the Cognitive Association Graph. This framework operates through a systematic, three-stage generation process. First, reasoning atoms are extracted from a curated seed set and assembled into the global graph; its vast combinatorial nature provides a near-infinite space of potential reasoning paths, directly enabling scalable generation. Next, a diversity-promoting random walk algorithm explores this structured space to sample long and intricate reasoning paths, the topological complexity of which forms the basis for problems of high difficulty. Finally, these paths are transformed by a constraint-based recombination mechanism, driven by three Cognitive Transfer Operators, which ensures the final combination is logically coherent while exhibiting high conceptual diversity.Our contributions are summarized as follows:

(1) We innovatively incorporate the established concept of reasoning atoms into a graph-based mathematical problem synthesis framework, enabling systematic extraction and representation of cognitive connections between fundamental mathematical concepts.

(2) We introduce three cognitive transition operators---Path Extension, Bridge Replacement, and Counterfactual Perturbation---that collectively ensure reasoning depth, logical coherence, and conceptual diversity in synthesized problems.

(3) Through extensive experimentation, we demonstrate that our CogAtom framework generates high-quality training data that significantly enhances foundation models' mathematical reasoning capabilities, with particularly pronounced improvements on advanced multi-step reasoning tasks.

\section{Related Work}

\subsection{Math Reasoning with LLMs}
While LLMs have demonstrated remarkable capabilities, their mathematical reasoning remains fragile, vulnerable to common failure modes such as distraction by irrelevant context \citep{yang2025how} and an inability to identify ill-defined problems with missing or contradictory conditions \citep{tian2024vcsearch}. To address these limitations, researchers have explored several approaches. Some work focuses on curating specialized datasets for math reasoning, aiming to offer more effective benchmarks and training resources for evaluating and enhancing model capabilities \citep{DBLP:conf/nips/HendrycksBKABTS21, DBLP:conf/nips/WangLXL24, DBLP:conf/iclr/YueQZFH00C24}. Another work leverages prompting strategies, such as chain-of-thought, to elicit more structured and accurate reasoning \citep{DBLP:conf/nips/Wei0SBIXCLZ22, DBLP:journals/corr/abs-2308-04371, DBLP:conf/iclr/FuPSCK23, DBLP:journals/corr/abs-2308-00304, DBLP:journals/corr/abs-2502-01694}. Beyond prompting, fine-tuning \citep{DBLP:journals/corr/abs-2309-02144,DBLP:journals/corr/abs-2409-12452,DBLP:journals/corr/abs-2502-03387},  in-context learning \citep{DBLP:conf/icml/ZhaoLK24}, reinforcement learning \citep{DBLP:journals/corr/abs-2311-09724, DBLP:conf/acl/WangLSXDLCWS24,DBLP:conf/acl/TrungZJSJL24}, test-time scaling \citep{DBLP:journals/corr/abs-2501-12948, DBLP:journals/corr/abs-2501-04519} and other strategies are also utilized to improve mathematical generalization and symbolic reasoning \citep{DBLP:journals/corr/abs-2411-14405, ma2025step, chen2025advancing,fu2025rlae}.

\subsection{Data Synthesis For Math Reasoning}
Early efforts in mathematical reasoning research often relied on manually constructed datasets \citep{DBLP:conf/nips/HendrycksBKABTS21, DBLP:journals/corr/abs-2110-14168, numinamath2024}. However, their limited scale and diversity have constrained the potential of LLMs in math reasoning tasks. Recent work has explored the use of LLMs themselves to generate synthetic data. Some work leverages LLMs to generate diverse problems through self-instruct, chain-of-thought prompting \citep{DBLP:conf/iclr/LuoSX0LTGLCT025,  DBLP:journals/corr/abs-2410-01560, DBLP:conf/iclr/YuJSYLZKLWL24, DBLP:conf/aaai/LiuZLY25}. Subsequent work introduces rejection sampling to alleviate the problem of low-quality data from direct prompting \citep{neal2003slice, DBLP:journals/corr/abs-2502-11476}. To move beyond simple prompting, recent work has focused on knowledge-driven synthesis pipelines. These approaches often extract structured elements, such as logically consistent templates \citep{huang2023solving} or key knowledge points \citep{DBLP:conf/icml/TangZWW24, DBLP:conf/aaai/HuangLGGSDC25}, from seed data to guide generation.

Although recent methods, especially knowledge-driven synthesis pipelines, have improved mathematical problem synthesis, they still suffer from shallow conceptual hierarchies and limited diversity. Furthermore, these approaches often apply a uniform generation budget, potentially overlooking the varied learning utility of problems at different difficulty levels \citep{xiong2025hsstar}. Our CogAtom framework addresses these challenges by introducing cognitive atoms and unique walk and recombination mechanism.

\section{Methodology}

\begin{figure*}[t!]
    \centering
    \includegraphics[width=\textwidth]{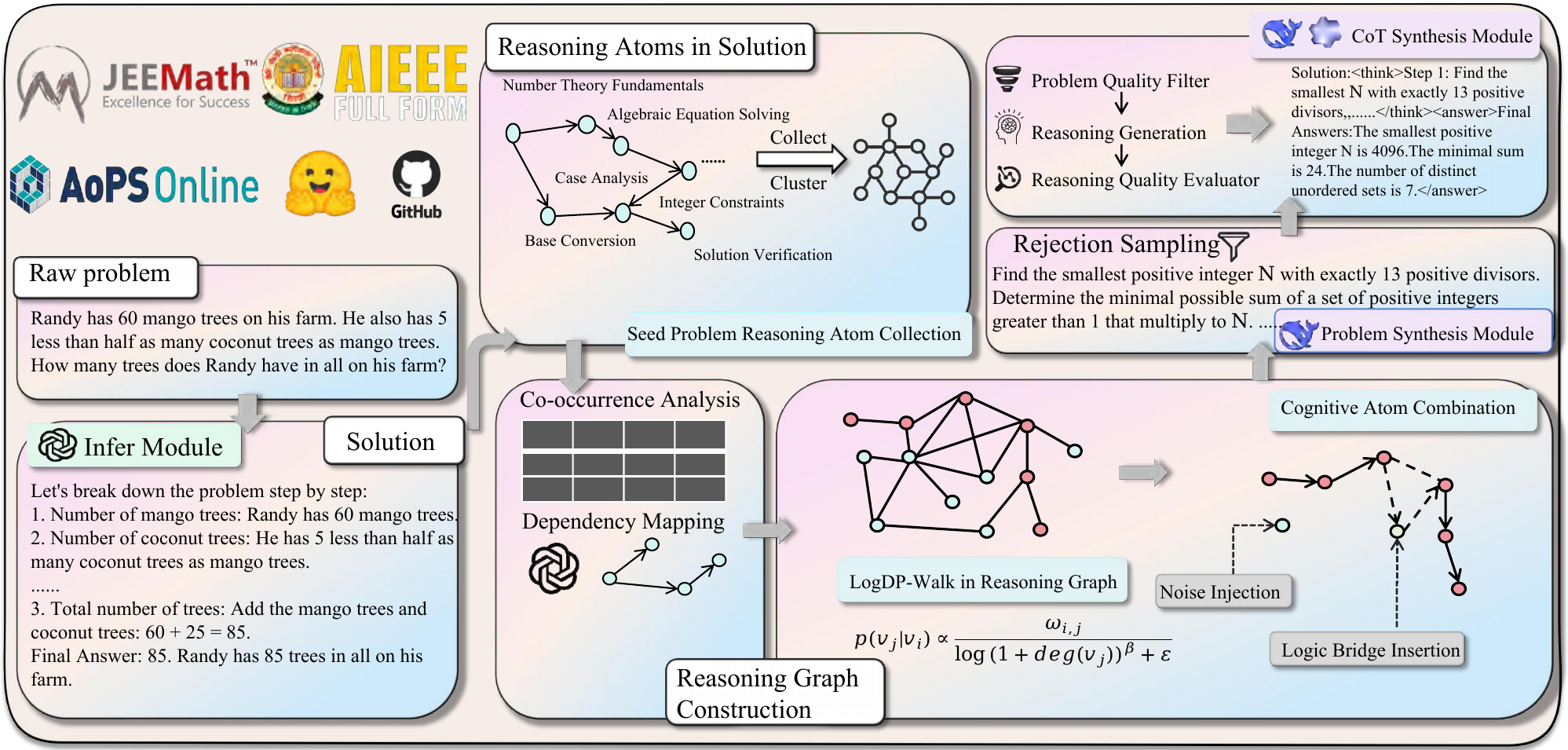} 
    \caption{The overall pipeline of CogAtom, which consists of three main stages. 
    \textbf{(1) Knowledge Base Construction:} Reasoning atoms are extracted from a curated seed set to build a global Cognitive Association Graph based on co-occurrence. 
    \textbf{(2) Reasoning Chain Generation:} A sample-and-refine process generates the final reasoning combination: diverse paths are first sampled from the global graph, then logically refined using local dependency information and Cognitive Transfer Operators.
    \textbf{(3) Problem Synthesis:} The refined combination serves as a logical blueprint to prompt a powerful LLM for the final synthesis of a new problem and its solution.}
    \label{fig:overview}
\end{figure*}

As shown in Figure~\ref{fig:overview}, our approach presents a comprehensive framework for mathematical problem synthesis, encompassing three crucial stages: (1) the extraction of reasoning atoms, (2) the construction of a cognitive association graph, and (3) the synthesis of challenging mathematical problems.

\subsection{Reasoning Atom Extraction}
\label{subsec:reasoning_atom_extraction}

The foundation of our framework is a rich and diverse set of reasoning atoms. The quality of these atoms is fundamentally constrained by the seed problems from which they are extracted. Guided by the principle that high-quality outputs necessitate high-quality inputs, we begin not with random data, but with a meticulously curated set of seed problems. To this end, mirroring efforts in computational education to automatically assess and filter high-quality simulated agents \citep{li2025exploring}, we introduce a systematic and reproducible procedure: the Automated Quality and Complexity Assessment Protocol.

This protocol leverages GPT-4o as an expert judge to evaluate candidate problems. First, we established a 5-point rubric to score problems based on the depth and complexity of reasoning required, inspired by established pedagogical principles and the design of large-scale educational datasets~ \citep{penedo2024fineweb}. The full rubric is detailed in Appendix~H. We then prompted GPT-4o to score each problem according to this rubric. To ensure robustness and mitigate potential biases, each problem was scored three times, and the average score was used. Finally, we applied a stringent filter, retaining only problems with an average score of 3.0 or higher. The rationale for this rigorous curation is twofold. First, by selecting problems that demand at least moderate multi-step or conceptual reasoning, we ensure that our initial pool of cognitive atoms is sufficiently rich and sophisticated to support the generation of novel, non-trivial problems. Second, this "quality-over-quantity" approach is supported by findings that an LLM's reasoning is more effectively unlocked by smaller, high-quality datasets \citep{DBLP:journals/corr/abs-2502-03387}, and that problems near the boundary of a model’s competence offer the most learning utility \citep{xiong2025hsstar}. Our protocol thus serves as a principled method for curating a high-quality set of problem-solving demonstrations.From this curated seed set of 9,403 problems, we then extract their constituent reasoning atoms. Inspired by AoT~\citep{DBLP:journals/corr/abs-2502-12018}, we prompt GPT-4o to solve each problem while reversely extracting the required atoms. To consolidate semantically redundant elements, we generate vector embeddings for the extracted atoms and cluster them based on cosine similarity, yielding a final, refined set of $|\mathcal{A}| = 44{,}117$ unique reasoning atoms, where $\mathcal{A}$ denotes the set of all reasoning atoms. 

\subsection{Graph-Based Reasoning Chain Generation}
\label{subsec:graph_and_chain_generation}

Our methodology transforms the curated set of reasoning atoms into novel and coherent problem structures through a sequential pipeline: (1) construction and refinement of a global knowledge graph to map conceptual associations, and (2) a sample-and-refine procedure on this graph to generate logical reasoning chains.

\paragraph{1. Global Graph Construction and Refinement.}
We begin by constructing a global, undirected Cognitive Association Graph $\mathcal{G}=(\mathcal{V}, \mathcal{E}, \omega)$, built on the principle of co-occurrence. This principle serves as a strong, data-driven proxy for the underlying cognitive associations in expert problem-solving. Each node $v \in \mathcal{V}$ is a unique reasoning atom. To mitigate the bias from high-frequency pairs, a common challenge in corpus-based network analysis, edges are weighted using a logarithmic transformation of their co-occurrence count: $\omega_{ij} = \log(1+n_{ij})$. To enhance the graph's utility for generating non-trivial problems, we then prune "supernodes"—nodes corresponding to overly generic concepts (e.g., "equation substitution"). These are identified statistically as nodes whose degree exceeds a threshold of $\mu+2\sigma$, where $\mu$ and $\sigma$ are the mean and standard deviation of node degrees in $G$. This process yields a pruned graph, $G'$, which serves as the foundational structure for exploration.

\paragraph{2. Reasoning Chain Generation.}
The generation of the final reasoning atom combinations is a two-stage process that moves from broad exploration to fine-grained logical refinement.

First, to generate diverse conceptual skeletons, we perform a biased random walk on the pruned graph $G'$, termed Diversity-Promoting Degree-Regularized Path Expansion (DPDRPE). At each step, the algorithm probabilistically selects the next node $v_{\text{next}}$ from the neighbors of the current node $v_{\text{curr}}$. The selection is governed by an association score that penalizes high-degree nodes, thus favoring less common and potentially more novel conceptual connections, defined as:
\begin{equation}
\label{eq:association_score}
\text{score}(v_{\text{next}}) = \frac{\omega_{v_{\text{curr}}, v_{\text{next}}}}{(\deg(v_{\text{next}}) + \epsilon)^\alpha}
\end{equation}
where $\omega$ is the co-occurrence weight, $\deg(\cdot)$ is the node degree, and $\alpha$ is a hyperparameter controlling the penalty strength. This yields a set of "reasoning paths" that are diverse but may lack strict logical coherence.

Second, to instill logical rigor, each sampled path undergoes an iterative refinement process formalized in Algorithm~\ref{alg:refinement}. The target combination size, $K$, is set to 10. This choice is informed by an analysis of human-authored Olympiad-level problems, which our analysis shows typically involve the synthesis of 8--12 core concepts. Our selection of $K=10$ thus aims to emulate a comparable level of cognitive complexity. For each path, we dynamically construct a local, directed Dependency Graph $G_{D,path} = (V_{path}, E_D)$, where $V_{path}$ contains the atoms in the sampled path $P$. Here, $s_{ij}$ represents the logical dependency strength between atoms $v_i$ and $v_j$ on a 5-point scale (detailed in Appendix~\ref{app:prompt_for_Cognitive_Atom_Extraction}). We retain only edges with $s_{ij} \geq 3$ (on a 5-point scale). This threshold acts as a crucial denoising step, filtering out weak or irrelevant connections while preserving meaningful, non-obvious relationships (scores 3 and above indicate moderate to strong logical dependencies) that are vital for creative synthesis. On this local graph, the conditional probability of a dependency is defined as $P(v_j|v_i) = s_{ij} / \sum_{v_k \in \text{succ}(v_i)} s_{ik}$, where $\text{succ}(v_i)$ is the set of successor nodes of $v_i$ in $G_{D,path}$.

\begin{algorithm}[t]
\caption{Reasoning Combination Refinement}
\label{alg:refinement}
\KwData{A sampled reasoning path $P$, Target size $K$}
\KwResult{A refined reasoning combination $C$}
$C \leftarrow \text{BackboneConstruction}(P)$ \tcp*{Select key nodes from P}
Construct local dependency graph $G_{D,path}$ for nodes in $C$\;
\While{$|C| < K$}{
    $C \leftarrow \text{BridgeReplacement}(C, G_{D,path})$\;
    $C \leftarrow \text{CounterfactualPerturbation}(C, P, G_{D,path})$\;
    $C \leftarrow \text{PathExtension}(C, G_{D,path})$\;
}
\Return{$C$}
\end{algorithm}
The refinement is driven by three Cognitive Transfer Operators, which are applied iteratively as outlined in Algorithm~\ref{alg:refinement}. Each operator is designed to optimize a specific property of the reasoning combination:

\noindent\textbf{Bridge Replacement} enhances logical coherence by inserting an intermediary node $v_k$ to connect a weakly linked pair $(v_i, v_j)$. The optimal bridge node is selected by maximizing the compound dependency strength, formalized as:
\begin{equation}
\label{eq:bridge_replacement}
v_k^* = \arg\max_{v_k \in V \setminus C} P(v_k|v_i) \cdot P(v_j|v_k)
\end{equation}

\noindent\textbf{Counterfactual Perturbation} promotes cognitive diversity by introducing an atom $v^*$ from the original path $P$ that is minimally associated with the current combination $C$. This encourages the exploration of novel conceptual links and is guided by:
\begin{equation}
\label{eq:counterfactual_perturbation}
v^* = \arg\min_{v \in P \setminus C} \max_{v_j \in C} P(v_j|v)
\end{equation}

\noindent\textbf{Path Extension} ensures the completeness and logical flow of the reasoning chain by appending a strongly dependent successor node $v_{next}$ to a node $v_i \in C$, governed by the condition: \begin{equation} \label{eq:path_extension} P(v_{next}|v_i) \geq \theta \end{equation} where $\theta$ is a predefined dependency threshold. Through this iterative process, diverse conceptual skeletons are transformed into combinations that are both logically sound and cognitively novel.Ultimately, we posit that the quality of a synthesized problem is directly determined by the logical coherence and conceptual novelty of its underlying reasoning chain—properties our sample-and-refine process is explicitly designed to optimize.

\subsection{Synthesis of Challenging Mathematical Problems}
Given a combination of reasoning atoms, we design an efficient pipeline for problem generation and quality control. Tailored prompts are crafted to guide large language models in synthesizing mathematical problems that are both challenging and diverse.To ensure the quality of the generated problems, we employ a rigorous multi-dimensional evaluation process that filters out questions lacking logical consistency, sufficient solvability, appropriate difficulty, or adequate concept coverage. For problems that pass this screening, we further utilize a strong teacher model to generate detailed step-by-step reasoning solutions. Each reasoning chain is then subjected to comprehensive quality assessment, focusing on conceptual integration, reasoning depth and rigor, key insight demonstration, error path exploration, and training applicability. Only those problems and reasoning chains that meet all quality criteria are retained in the final dataset.Through this dual-stage quality assurance process, we construct a dataset $\mathcal{D} = \{(q_i, s_i, a_i)\}$, where $q_i$ denotes the problem statement, $s_i$ is the step-by-step solution, and $a_i$ is the final answer. This dataset provides a solid foundation for training and evaluating advanced mathematical reasoning models.

\section{Experiments}

\subsection{Datasets}

\textbf{Seed Dataset Construction.} To establish a high-quality foundation for our synthetic data generation process, we constructed a comprehensive seed dataset comprising 9,403 mathematical problems carefully selected from multiple established datasets. These source datasets include: GSM8K \citep{DBLP:journals/corr/abs-2110-14168} (MIT License), MATH \citep{DBLP:conf/nips/HendrycksBKABTS21} (MIT license), TAL-SCQ5K-EN \citep{matheval2023} (MIT License), JEEE and AIEEE\footnote{https://jeemath.in/}, ranging from elementary and middle school difficulty to Olympic-level difficulty. To ensure quality and appropriate difficulty distribution, we employed GPT-4o \citep{DBLP:journals/corr/abs-2405-00732} to filter and categorize the problems based on their complexity and reasoning requirements, resulting in a diverse and balanced seed collection suitable for our synthetic data generation pipeline. The detailed composition of the seed dataset is summarized in Table~\ref{tab:seed_data_composition}.

\begin{table}[htbp]
    \centering
    \small 
    \setlength{\tabcolsep}{12pt}{
    \begin{tabular}{l r}
        \toprule
        \textbf{Source Dataset} & \textbf{Number of Samples} \\
        \midrule
        GSM8K (train) & 1351 \\
        TAL-SCQ5K-EN (train) & 526 \\
        MATH (train) & 3994 \\
        AIEEE/JEEE & 3472 \\
        \midrule
        \textbf{Total} & \textbf{9343} \\
        \bottomrule
        \end{tabular}}
    \caption{Composition of the seed dataset used for synthetic data generation.}
    \label{tab:seed_data_composition}

\end{table}

\subsection{Baseline Methods}

Our evaluation employs both short-CoT (concise intermediate reasoning steps) and long-CoT (extended reasoning with self-reflection and alternative paths) approaches. We compare against five state-of-the-art mathematical problem generation methods:

For comprehensive assessment, we compare our method against several prominent approaches in mathematical problem generation: \textbf{Evol-Instruct} \citep{DBLP:conf/iclr/LuoSX0LTGLCT025} implements an iterative refinement mechanism using LLMs to progressively enhance instructional data complexity; \textbf{KPDDS} \citep{DBLP:conf/aaai/HuangLGGSDC25} extracts key points from authentic sources to generate mathematically coherent question-answer pairs; \textbf{OpenMath} \citep{DBLP:journals/corr/abs-2410-01560} synthesizes solutions for established benchmarks through open-source language models; \textbf{NuminaMath} \citep{numinamath2024} provides competition-level mathematical problems with detailed reasoning traces; and \textbf{MathScale} \citep{DBLP:conf/icml/TangZWW24} constructs concept graphs from seed questions to guide diverse problem generation.

For methods without publicly released problem sets (specifically Evol-Instruct and KPDDS), we followed their documented methodologies using Qwen2.5-Math-72B-Instruct \citep{DBLP:journals/corr/abs-2409-12122} to generate comparable problem collections. For NuminaMath, OpenMathInstruct, and MathScale, we utilized their published problem sets directly.

\subsection{Implementation Details}
We employ GPT-4o to generate step-by-step solutions, from which atomic reasoning steps were extracted. Each reasoning atom is encoded as a dense vector using the BGE-M3 model \citep{DBLP:journals/corr/abs-2402-03216}, followed by L2 normalization. We apply MiniBatch KMeans clustering to these embeddings and further merged highly similar clusters using cosine similarity filtering. We then construct a cognitive association graph comprising 44,177 reasoning atom nodes and 149,576 edges. To generate diverse reasoning paths, we perform iterative degree-penalized random walks of order $n=5$ starting from each node in the concept graph, thereby constructing cognitive reasoning paths. Along each path, we apply three types of cognitive leap operations to obtain combinations of reasoning atoms, with each combination containing 10 nodes. For problem synthesis, we use Qwen2.5-72B-Instruct to generate CogAtom-short problems and DeepSeek-R1-Distill-Qwen-32B \citep{DBLP:journals/corr/abs-2501-12948} for CogAtom-long problems. For quality control, we employ Qwen2.5-72B-Instruct as an LLM-based judge. We further fine-tune Qwen2.5-Math-7B \citep{DBLP:journals/corr/abs-2409-12122} and Qwen2.5-14B-Origin using the Adam optimizer with initial learning rates of $2 \times 10^{-5}$ and $1 \times 10^{-5}$, respectively. To further investigate the reasoning ability of long-CoT, we fine-tuned DeepSeek-R1-Distill-Qwen-7B and evaluate on three high-difficulty datasets (MATH, AIME 2024 and AIME 2025). All training was conducted with BF16 mixed precision and Flash Attention for two epochs, and greedy decoding is used during evaluation. All experiments were conducted on a cluster of 8 machines, each equipped with NVIDIA A100 GPUs.

\subsection{Main Result}
\begin{table*}[]
\centering
\small 
\setlength{\tabcolsep}{20pt}{
\begin{tabular}{@{}lccccc@{}}
\toprule

\textbf{Methods} & \textbf{gsm8k} & \textbf{math500} & \textbf{AMC} & \textbf{AIME2024} & \textbf{AIME2025} \\ \midrule
\multicolumn{6}{c}{\textit{Models based on Qwen2.5-Math-7B}} \\ 
\midrule
KPDDS                               & 89.5\% & 73.6\% & \underline{50.0\%} &  \underline{4/30} &  \underline{2/30} \\ 
Evol-Instruct                       & 87.9\% & 73.8\% & 45.0\% & 2/30 & 1/30 \\ 
NuminaMath                          & 84.4\% & 70.6\% & 47.5\% & 2/30 & 1/30 \\
MathScale                           & 80.8\% & 71.2\% & 45.0\% & 1/30 &  \underline{2/30} \\
OpenMath                            & 87.2\% & 69.0\% & 47.5\% & 1/30 &  \underline{2/30} \\ 
CogAtom-short              & \underline{90.4\%} & \underline{75.0\%} & \textbf{52.5\%} & 3/30 &  \underline{2/30} \\ 
CogAtom-long               & \textbf{91.2\%} & \textbf{83.0\%} & 47.5\% & \textbf{5/30} & \textbf{3/30} \\ \midrule

\multicolumn{6}{c}{\textit{Models based on Qwen2.5-14B-Origin}} \\ 
\midrule
KPDDS                               & 88.8\% & 58.8\% & 29.1\% & \textbf{3/30} &  \underline{1/30} \\ 
Evol-Instruct                       & 87.8\% & 63.0\% & 32.5\% &  \underline{2/30} &  \underline{1/30} \\
NuminaMath                          & 79.0\% & 59.0\% & 37.5\% & 1/30 &  \underline{1/30} \\
MathScale                           & 80.6\% & 59.4\% & 22.5\% & 1/30 & 0/30 \\
OpenMath                            & 87.8\% & 64.2\% &  \underline{40.0\%} &  \underline{2/30} &  \underline{1/30} \\ 
CogAtom-short                       &  \underline{89.2\%} &  \underline{68.8\%} & \textbf{45.0\%} & \textbf{3/30} & \textbf{2/30} \\
CogAtom-long                        & \textbf{91.7\%} & \textbf{76.4\%} & 32.5\% & \textbf{3/30} & \textbf{2/30} \\ \bottomrule
\end{tabular}}
\caption{Evaluation results on five mathematical benchmarks for model Qwen2.5-Math-7B and Qwen2.5-14B-Origin, both fine-tuning with 100K synthetic problems. Within each section, the best results are highlighted in bold font, and the second best results are underlined. The number of correct answers (out of 30) is reported for both AIME2024 and AIME2025.}
\label{tab: main-result-qwen}
\end{table*}

Tables \ref{tab: main-result-qwen} and~\ref{tab:main-result-dsr1} present comprehensive evaluation results comparing our CogAtom framework against state-of-the-art mathematical problem generation methods across five benchmarks of increasing difficulty. Our analysis reveals several significant findings:

(1) When used for fine-tuning identical base models, CogAtom-generated data consistently yields superior performance across all benchmarks. For Qwen2.5-Math-7B, fine-tuning with CogAtom-long data achieves 91.2\% accuracy on GSM8K and 83.0\% on MATH500, outperforming the next best data sources (KPDDS at 89.5\% and Evol-Instruct at 73.8\%, respectively). Similar advantages are observed with Qwen2.5-14B-Origin, demonstrating the cross-architectural transferability of our approach.

(2) The performance advantage from CogAtom-generated training data becomes increasingly pronounced as problem difficulty increases. With Qwen2.5-Math-7B, the improvement margin grows from 1.7 percentage points on GSM8K to 9.2 percentage points on MATH500 (compared to the strongest baselines). For Olympic-level problems, Qwen2.5-Math-7B fine-tuned on CogAtom-long data correctly solves $5/30$ AIME2024 problems, compared to $4/30$ for the best baseline method. This pattern highlights our framework's effectiveness in generating training data that encodes complex reasoning patterns required for advanced mathematical problem-solving.

(3) Training with CogAtom-short data yields models that excel on structured problems with clear solution paths (e.g., 52.5\% on AMC using Qwen2.5-Math-7B), while CogAtom-long data produces models that perform better on problems requiring multi-step reasoning (e.g., 83.0\% vs. 75.0\% on MATH500). This differentiation reflects how our cognitive leap operators create training examples that develop distinct reasoning capabilities based on the complexity of target tasks.

(4) As shown in Table~\ref{tab:main-result-dsr1}, when fine-tuning DeepSeek-R1-Distill-Qwen-7B, CogAtom-long data enables achieving 90.8\% accuracy on MATH500, surpassing the next best method (Evol-Instruct at 79.6\%) by 11.2 percentage points. Most remarkably, it facilitates solving $10/30$ AIME2024 problems and $9/30$ AIME2025 problems. These substantial improvements demonstrate that training examples generated by CogAtom capture complex conceptual dependencies and multi-step reasoning paths, thereby unlocking the full potential of advanced reasoning models.

\begin{table}[t!]
\centering
\small 
\setlength{\tabcolsep}{6pt}
\begin{tabular}{@{}lccc@{}}
\toprule
\multicolumn{1}{l}{\textbf{Methods}} & \multicolumn{1}{c}{\textbf{math500}} & \multicolumn{1}{c}{\textbf{AIME2024}} & \multicolumn{1}{c}{\textbf{AIME2025}} \\ 
\midrule
KPDDS & 76.6\% & \underline{6/30} & 1/30 \\
Evol-Instruct & \underline{79.6\%} & \underline{6/30} & \underline{2/30} \\ 
NuminaMath & 72.2\% & 4/30 & \underline{2/30} \\
MathScale & 75.3\% & 1/30 & 0/30 \\
OpenMath & 68.0\% & 2/30 & \underline{2/30} \\
CogAtom-long & \textbf{90.8\%} & \textbf{10/30} & \textbf{9/30} \\ 
\bottomrule
\end{tabular}
\caption{Evaluation results on three mathematical benchmarks with high-difficulty for model Qwen2.5-7B-DeepSeek-R1-Distill-Qwen-7B.}
\label{tab:main-result-dsr1}
\end{table}

\subsection{Analysis of Data Scale}
\label{subsec:math_scaling}

\begin{figure}[t!]
\centering
\includegraphics[width=\columnwidth]{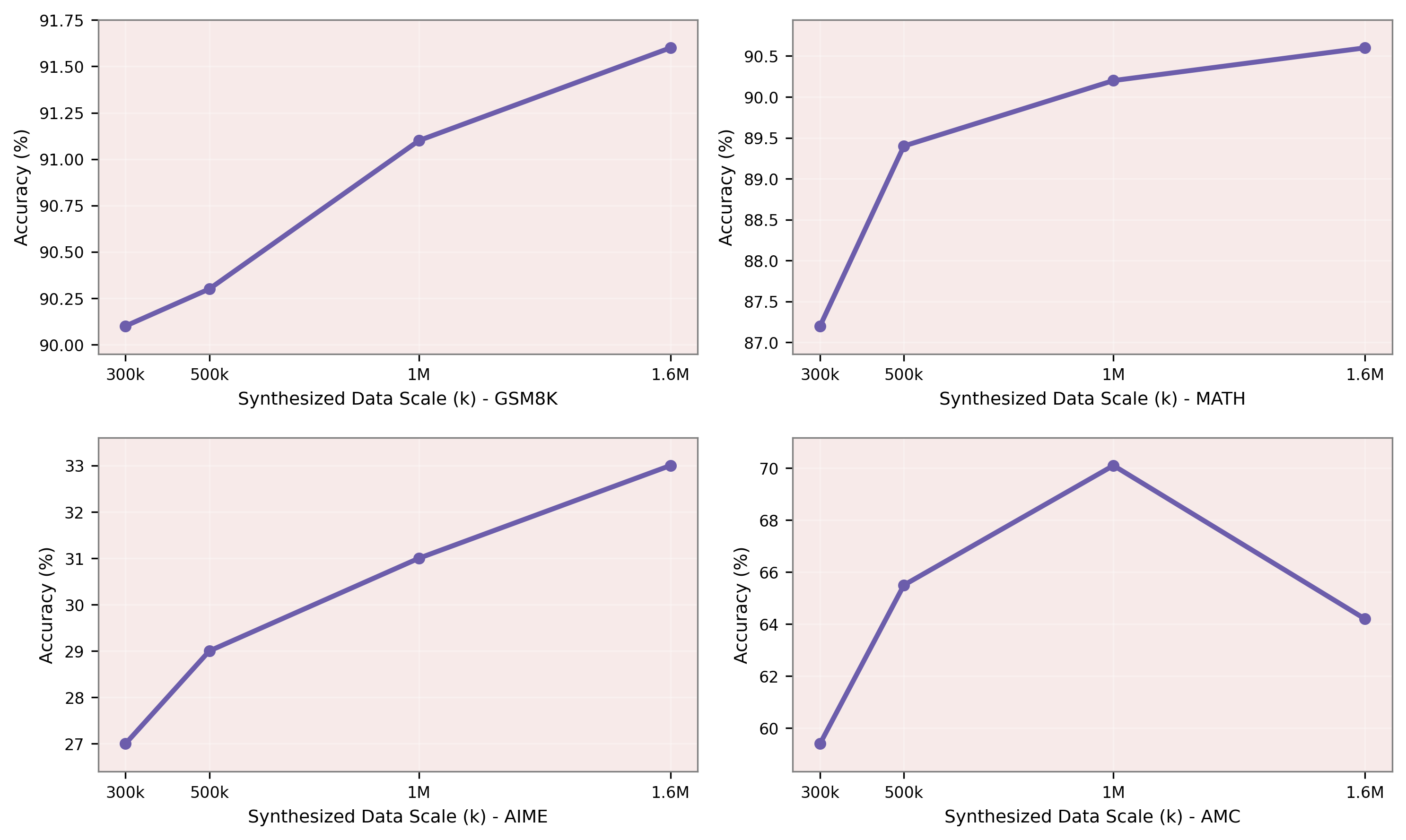}
\caption{Performance on mathematics benchmarks with increasing scale of synthesized data from CogAtom-long. We report accuracy for GSM8K, MATH, AIME, and AMC.}
\label{fig:math_scaling}
\end{figure}

To evaluate the scalability of our data synthesis engine, we fine-tuned Qwen2.5-7B-Math on CogAtom-generated datasets of increasing sizes, from 300k up to 1.6 million problems. The results, presented in Figure~\ref{fig:math_scaling} and detailed in Table~\ref{tab:scaling_details}, reveal a strong, positive correlation between data scale and performance on core mathematical reasoning benchmarks.

The scaling trend is particularly pronounced on the most challenging benchmarks. On the competition-level AIME dataset, for instance, accuracy exhibits a clear monotonic ascent with data volume: it rises from 27.0\% at 300k samples, to 29.0\% (+500k), 31.0\% (+1M), and culminates at 33.0\% with the full 1.6M dataset. This scaling behavior, characterized by diminishing returns, is consistent with established logarithmic scaling laws in large language models. Furthermore, the data reveals a differentiated impact: the absolute performance gain on the complex reasoning required for AIME (+6.0\% from 300k to 1.6M) is substantially larger than on the more algorithmic GSM8K benchmark (+1.5\% over the same interval). While performance on AMC shows a different trajectory, peaking at 1M samples, the overall results strongly validate that the CogAtom engine is a scalable and effective method for enhancing advanced mathematical reasoning.

\begin{table}[t!]
\centering
\resizebox{\columnwidth}{!}{%
\begin{tabular}{@{}lcccc@{}}
\toprule
\textbf{Data Scale} & \textbf{GSM8K} & \textbf{MATH} & \textbf{AIME} & \textbf{AMC} \\
\midrule
+300k  & 90.1\% & 87.2\% & 27.0\% & 59.4\% \\
+500k  & 90.3\% & 89.4\% & 29.0\% & 65.5\% \\
+1M    & 91.1\% & 90.2\% & 31.0\% & 70.1\% \\
+1.6M  & 91.6\% & 90.6\% & 33.0\% & 64.2\% \\
\bottomrule
\end{tabular}%
}
\caption{Detailed performance on mathematics benchmarks with increasing scale of synthesized data from CogAtom-long.}
\label{tab:scaling_details}
\end{table}

\subsection{Cross-Domain Generalization to Physics}
\label{subsec:physics_generalization}
\begin{table*}[ht]
\centering
\small
\setlength{\tabcolsep}{4pt}
\caption{Performance on physics benchmarks as the scale of synthetic data increases. We report accuracy scores. The best results are in bold.}
\label{tab:physics_generalization}
\begin{tabular}{@{}lccccc@{}}
\toprule
\textbf{Synthetic Data Scale} & \textbf{C-MMLU} & \textbf{C-MMLU} & \textbf{MMLU} & \textbf{MMLU} & \textbf{MMLU} \\
& Concept. Physics & High School Physics & Concept. Physics & High School Physics & Univ. Physics \\
\midrule
Baseline (0 samples) & 0.7687 & 0.6818 & 0.7050 & 0.5762 & 0.5000 \\
+300k (Our Method) & 0.7823 & 0.6909 & 0.7450 & 0.5960 & \textbf{0.5882} \\
+600k (Our Method) & \textbf{0.7959} & \textbf{0.7455} & 0.7200 & \textbf{0.6093} & 0.5294 \\
\bottomrule
\end{tabular}
\end{table*}

To assess the domain-agnostic nature of the CogAtom paradigm, we applied it to physics—a domain that, like mathematics, is characterized by complex principles and multi-step reasoning.We synthesized two large-scale datasets of 300k and 600k physics problems, respectively, and used them to fine-tune the Qwen2.5-7B base model.The empirical results, presented in Table~\ref{tab:physics_generalization}, provide strong support for this hypothesis. Fine-tuning on the generated data yields consistent and substantial performance gains across a suite of standard physics benchmarks in both Chinese (C-MMLU) and English (MMLU). The performance improvements scale positively with data volume; for instance, on C-MMLU High School Physics, the accuracy gain grows from +1.3\% with 300k samples to a remarkable \textbf{+9.3\%} with 600k samples. This monotonic trend underscores the effectiveness of our synthesized data in imparting robust physical reasoning skills. A qualitative case study of a generated physics problem, which integrates concepts from thermodynamics and black-body radiation, is detailed in Appendix~\ref{app:physics_case_study_analysis}. 
\begin{table*}[ht]
\centering
\small
\setlength{\tabcolsep}{18pt}
\begin{tabular}{@{}llcccc@{}}
\toprule
\textbf{Methods} & \textbf{Ablation} & \textbf{gsm8k} & \textbf{math500} & \textbf{AIME2024} & \textbf{AIME2025} \\
\midrule
\multirow{4}{*}{CogAtom-short}
    & Full model                 & \textbf{88.5\%} & \textbf{70.6\%} & \textbf{3/30} & \textbf{2/30} \\
    & w/o degree-penalty         & 87.1\%          & 69.6\%          & \underline{2/30} & \underline{1/30} \\
    & w/o cognitive              & \underline{87.6\%} & \underline{70.2\%} & \textbf{3/30} & \textbf{2/30} \\
    & w/o reject-sampling        & 86.9\%          & 70.6\%          & 1/30 & \textbf{2/30} \\
\midrule
\multirow{4}{*}{CogAtom-long}
    & Full model                 & \textbf{91.0\%} & \textbf{70.6\%} & \textbf{4/30} & \textbf{3/30} \\
    & w/o degree-penalty         & 90.3\%          & \underline{69.2\%} & 1/30 & \underline{2/30} \\
    & w/o cognitive              & \underline{90.6\%} & 68.0\%          & 2/30 & \underline{2/30} \\
    & w/o reject-sampling        & 89.5\%          & 69.0\%          & \underline{3/30} & \underline{2/30} \\
\bottomrule
\end{tabular}
\caption{Ablation study results of different synthesis components for model Qwen2.5-Math-7B finetuning with 10K synthetic problems.}
\label{tab:ablation-components}
\end{table*}
\subsection{Ablation Study on Synthesis Components}

Table~\ref{tab:ablation-components} reports the results of our ablation study on the key components of the framework. The degree-penalty mechanism is crucial for promoting conceptual diversity by mitigating the selection bias toward high-degree nodes; its removal causes a dramatic drop in CogAtom-long's performance on AIME2024 from $4/30$ to $1/30$, especially for complex problems. Cognitive leap operators are particularly beneficial for long-form reasoning: their ablation leads to a $2.6$ percentage point decrease in MATH500 accuracy and halves the AIME2024 score for CogAtom-long. In contrast, these operators have minimal effect on CogAtom-short, indicating their primary role in enhancing complex reasoning chains. Quality-based rejection sampling also plays a significant role in challenging benchmarks; without it, CogAtom-short's AIME2024 result declines from $3/30$ to $1/30$. Collectively, these findings confirm that each component makes a meaningful contribution, with their importance varying according to the complexity of mathematical reasoning tasks.

\subsection{Analysis of Problem Difficulty}

We assess problem difficulty using two complementary metrics: answer consistency and inference tokens. A problem is marked as consistent if both models generate identical answers. We also record the total number of tokens consumed during their reasoning to assess difficulty. We additionally extend our analysis to the challenging Olympiad-level AIME2024 dataset. The results are presented in Table \ref{tab:difficulty}. Compared with other baseline synthetic methods, our CogAtom framework achieving the lowest answer consistency (67.47\%) and most tokens (3022), indicating synthesis of the most challenging problem. Although AIME2024 benchmark still yields an even lower consistency of 50.00\% and more tokens of 5260, CogAtom narrows this gap more than any other methods, demonstrating its effectiveness at generating higher-difficulty, more realistic mathematical challenges.

\begin{table}[ht]
\centering
\small
\begin{tabular}{@{}lcc@{}}
\toprule
\textbf{Methods}    & \textbf{Answer Consistency} & \textbf{Tokens} \\ \midrule
KPDDS              & 75.7\%                        & 2328            \\
Evol-Instruct      & 70.0\%                       & 2282            \\
NuminaMath         & 86.3\%                        & 1954            \\
MathScale          & 68.6\%                        & 2103            \\
OpenMath           & 80.5\%                        & 2532            \\
AIME2024           & \textbf{50.0\%}               & \textbf{5260}            \\
CogAtom            & \underline{67.5\%}            & \underline{3022}            \\ \bottomrule
\end{tabular}
\caption{Analysis of problem difficulty with model Qwen2.5-72B and DeepSeek-R1-distill-Qwen-32B.}
\label{tab:difficulty}
\end{table}

\subsection{Analysis of Problem Diversity}
\begin{table}[ht]
\centering
\small
\setlength{\tabcolsep}{24pt}{
\begin{tabular}{@{}lc}
\toprule
\textbf{Methods}    & \textbf{PTD}\\ \midrule
KPDDS              & 1.7903\\
Evol-Instruct      & 1.7931\\
NuminaMath         & \underline{1.7936}\\
MathScale          & 1.7836\\
OpenMath           & 1.7190\\
AIME2024           & 1.7896\\
CogAtom            & \textbf{1.7961}\\ \bottomrule
\end{tabular}}
\caption{Analysis of problem diversity}
\label{tab:diversity}
\end{table}
We further analyze the diversity of synthesized problems. We introduce the Problem Type Diversity (PTD) metric to rigorously quantify semantic diversity in mathematical problem datasets. PTD jointly captures the breadth of problem type coverage and the uniformity of their distribution:
\begin{equation}
\text{PTD} = \frac{N_c}{\sqrt{N}} \left(1 - \frac{\sigma_c}{\mu_c \sqrt{N_c}}\right)
\end{equation}
Here, $N_c$ is the semantic cluster count, $N$ is the total sample size, $\mu_c$ is the mean cluster size, and $\sigma_c$ is the cluster size standard deviation. Empirically, CogAtom-Long achieves the highest PTD score (1.7961), reflecting comprehensive and balanced problem coverage. In contrast, baseline datasets like KPDDS (1.7190) show lower PTD, indicating more homogeneous content. These results highlight our approach's advantage in fostering semantic diversity, critical for enhancing model generalization to novel mathematical reasoning tasks.

\section{Conclusion}
In this paper, we introduced CogAtom, a novel framework for mathematical problem synthesis that integrates reasoning atoms and cognitive association graphs to generate high-quality training data. Our extensive experiments demonstrated that models fine-tuned on CogAtom-generated problems achieve substantial improvements in mathematical reasoning capabilities, particularly on advanced multi-step reasoning tasks, outperforming existing methods by significant margins on challenging benchmarks including MATH500 and AIME. The effectiveness of our approach highlights the importance of cognitive science principles in designing synthetic training data for enhancing reasoning abilities in foundation models.

\section*{Limitations}
Despite the promising results of our work, a primary limitation is that the CogAtom framework currently operates exclusively in the textual modality. This constraint limits its applicability to mathematical domains that are inherently visual, such as geometry and graph theory, where diagrams and figures are often integral to the problem statement. A significant direction for future research is to extend CogAtom into a multimodal framework. This would involve developing methods to represent visual components as a new type of cognitive atom and learning the cross-modal associations between textual and visual elements. Such an extension would enable the synthesis of a far richer and more comprehensive class of problems, better reflecting the multimodal nature of human mathematical reasoning.

\section*{Ethics Statement}
All experiments in this study are conducted on publicly available and commonly used datasets. We acknowledge the risks associated with automated content generation by large language models. Our research on structured data synthesis contributes to the development of more controllable and reliable methodologies, aiming to foster beneficial applications of AI in specialized reasoning domains.

\section*{Acknowledgements}

This work was supported in part by the National Natural Science Foundation of China under Grant 62477001.

\bibliography{main}

\appendix

\section{Seed Problem Curation Strategy}
\label{app:curation_strategy}

To ensure our seed problem set possesses sufficient complexity and quality to foster advanced reasoning, we developed and implemented a systematic curation strategy. Our approach is grounded in the principle that the quality of synthesized data is fundamentally dependent on the quality of the initial seeds. This strategy automates the curation process in a principled and reproducible manner.

\begin{figure*}[t!]
\centering
\includegraphics[width=1\textwidth]{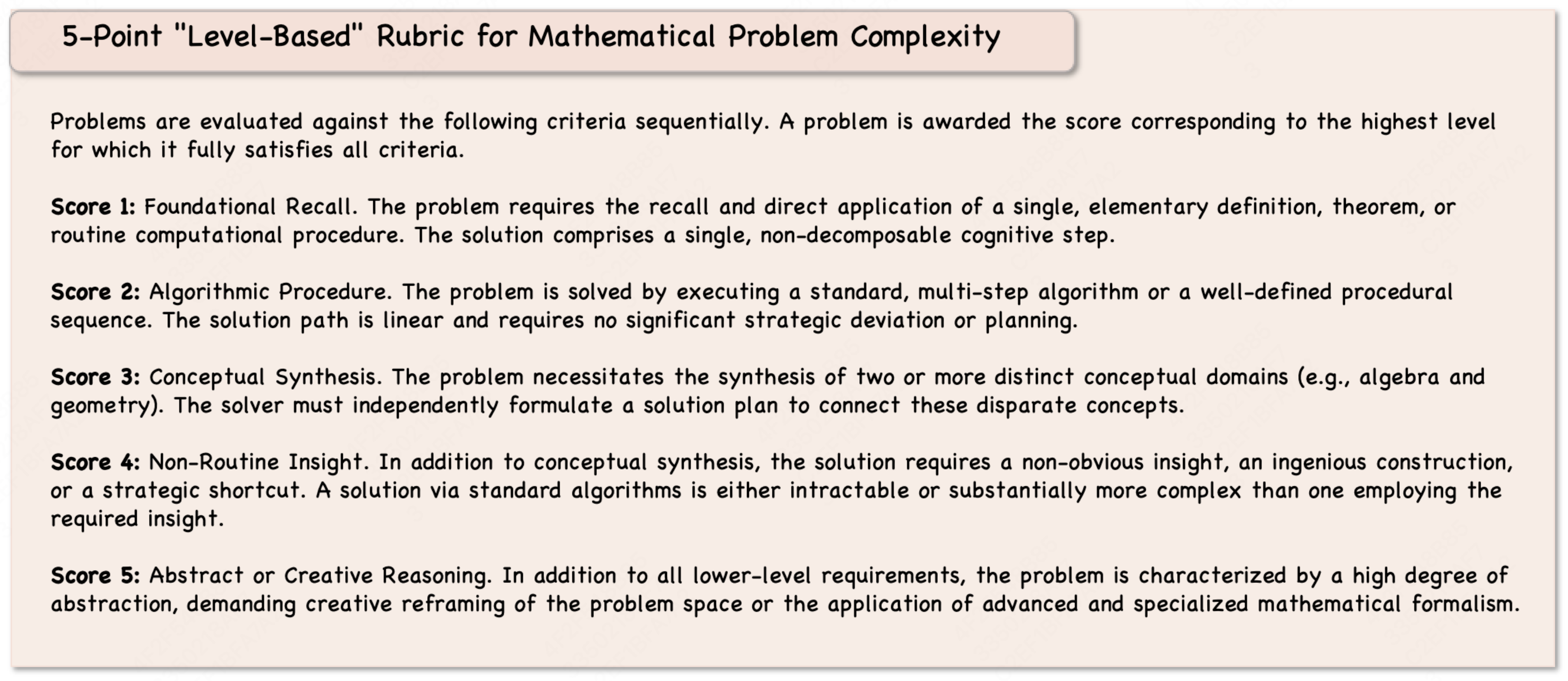}
\caption{The 5-level, level-based rubric used in our seed problem curation strategy. Each level defines a specific set of cumulative criteria for assessing the reasoning depth of a mathematical problem, from Foundational Recall (Score 1) to Abstract or Creative Reasoning (Score 5).}
\label{fig:complexity_rubric}
\end{figure*}

The core of this strategy is an LLM-as-an-expert-judge assessment. For each candidate problem, we prompted GPT-4o to score it three independent times to ensure robustness and mitigate potential scoring anomalies. The final score was the average of these three ratings. Subsequently, we applied a stringent quality filter, retaining only problems with an average score of 3.0 or higher. The evaluation rubric used in this process, which provides a structured hierarchy for assessment from foundational recall to abstract reasoning, is presented in Figure~\ref{fig:complexity_rubric}.

\section{Prompt for Cognitive Atom Extraction}
\label{app:prompt_for_Cognitive_Atom_Extraction}
\begin{figure*}
    \includegraphics[width=1\linewidth]{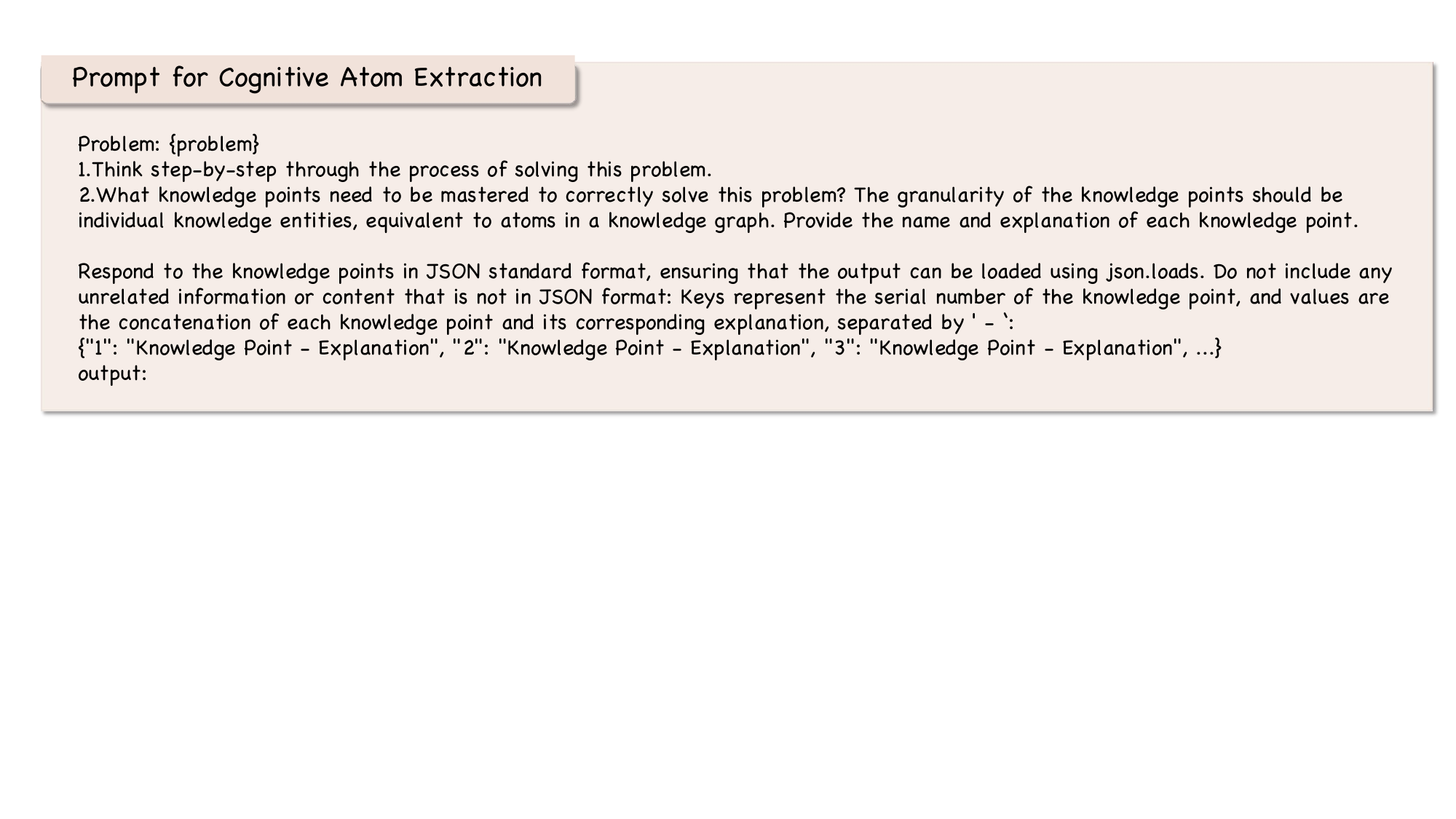}
    \caption{Prompt for extracting cognitive atom from problems}
    \label{fig:cog-atom-extraction-prompt}
\end{figure*}

Figure \ref{fig:cog-atom-extraction-prompt} instructs the language model to analyze mathematical problems through a step-by-step reasoning process, then identify and extract the fundamental cognitive atoms—atomic knowledge entities at appropriate granularity—required to master the problem, enabling systematic representation of the core mathematical concepts and principles underlying complex reasoning tasks.

\begin{figure*}
    \includegraphics[width=1\linewidth]{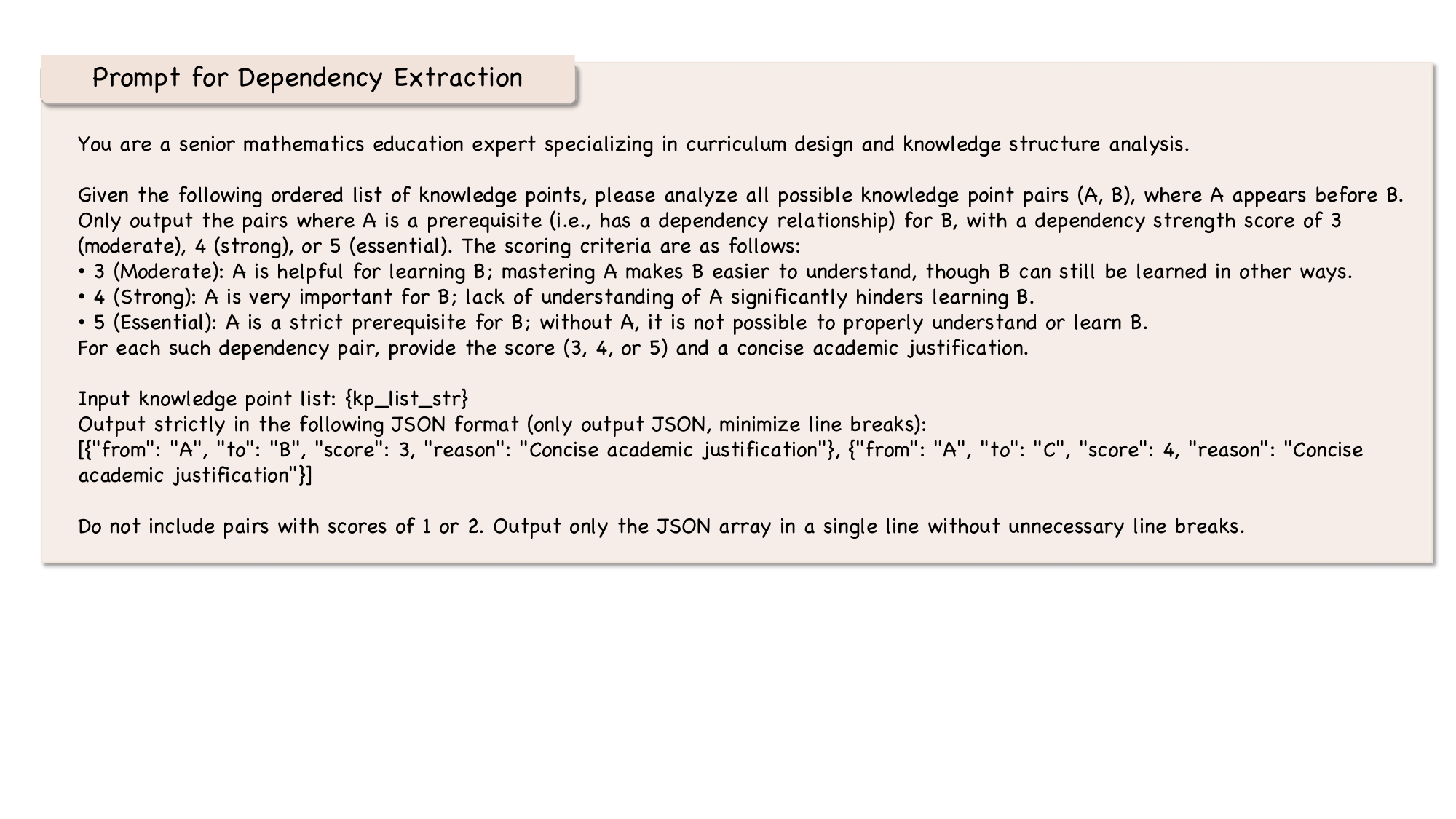}
    \caption{Prompt for extracting dependency}
    \label{fig:dependency-extraction-prompt}
\end{figure*}

Figure \ref{fig:dependency-extraction-prompt} presents the prompt used to quantify logical dependencies between cognitive atoms. This prompt guides the language model to evaluate the strength of prerequisite relationships on a 5-point scale, enabling the construction of dependency graphs for reasoning chain refinement.

\begin{figure*}
    \includegraphics[width=1\linewidth]{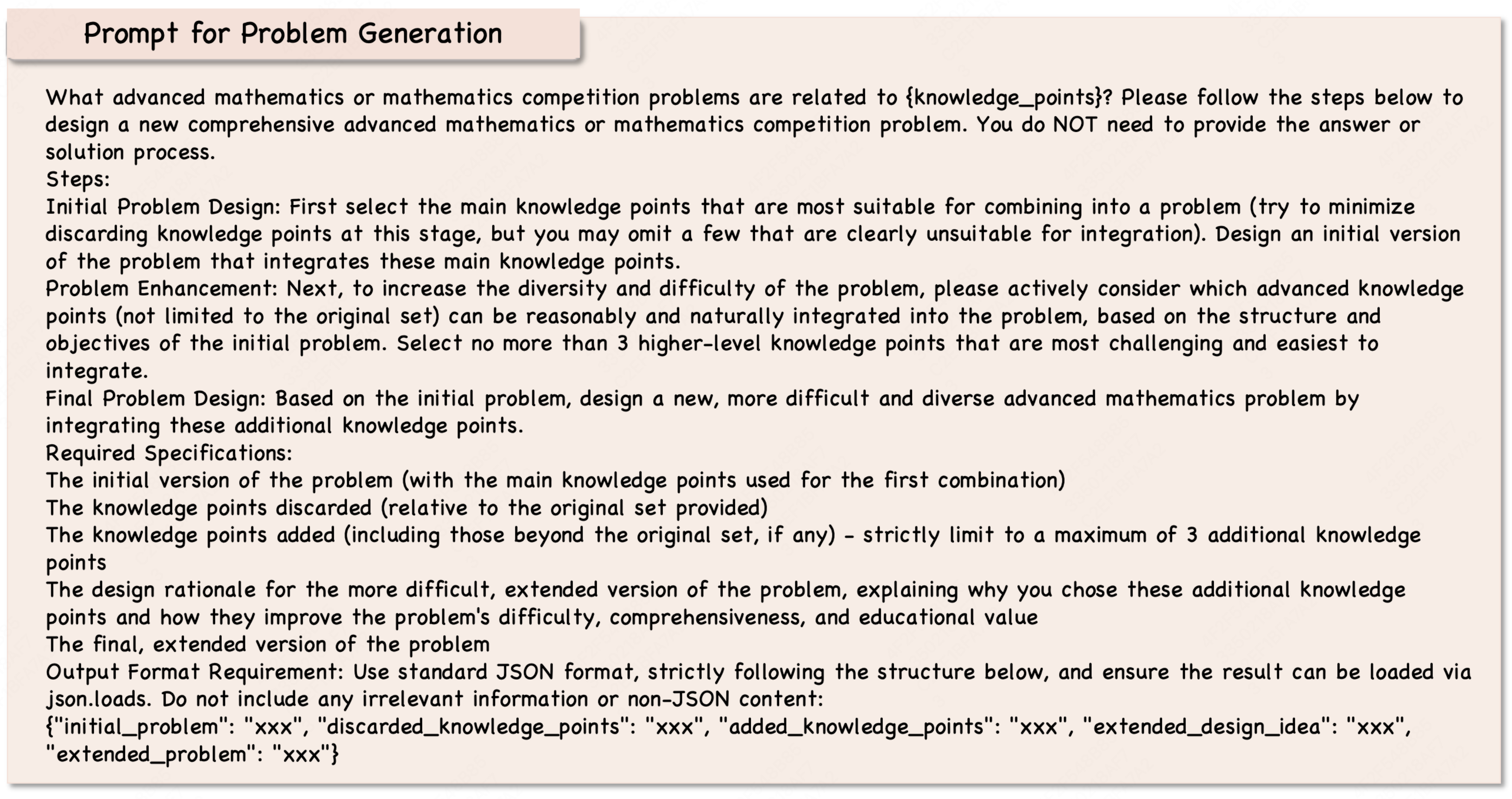}
    \caption{Prompt for problem generation}
    \label{fig:problem-generation-prompt}
\end{figure*}

\begin{figure*}
    \includegraphics[width=1\linewidth]{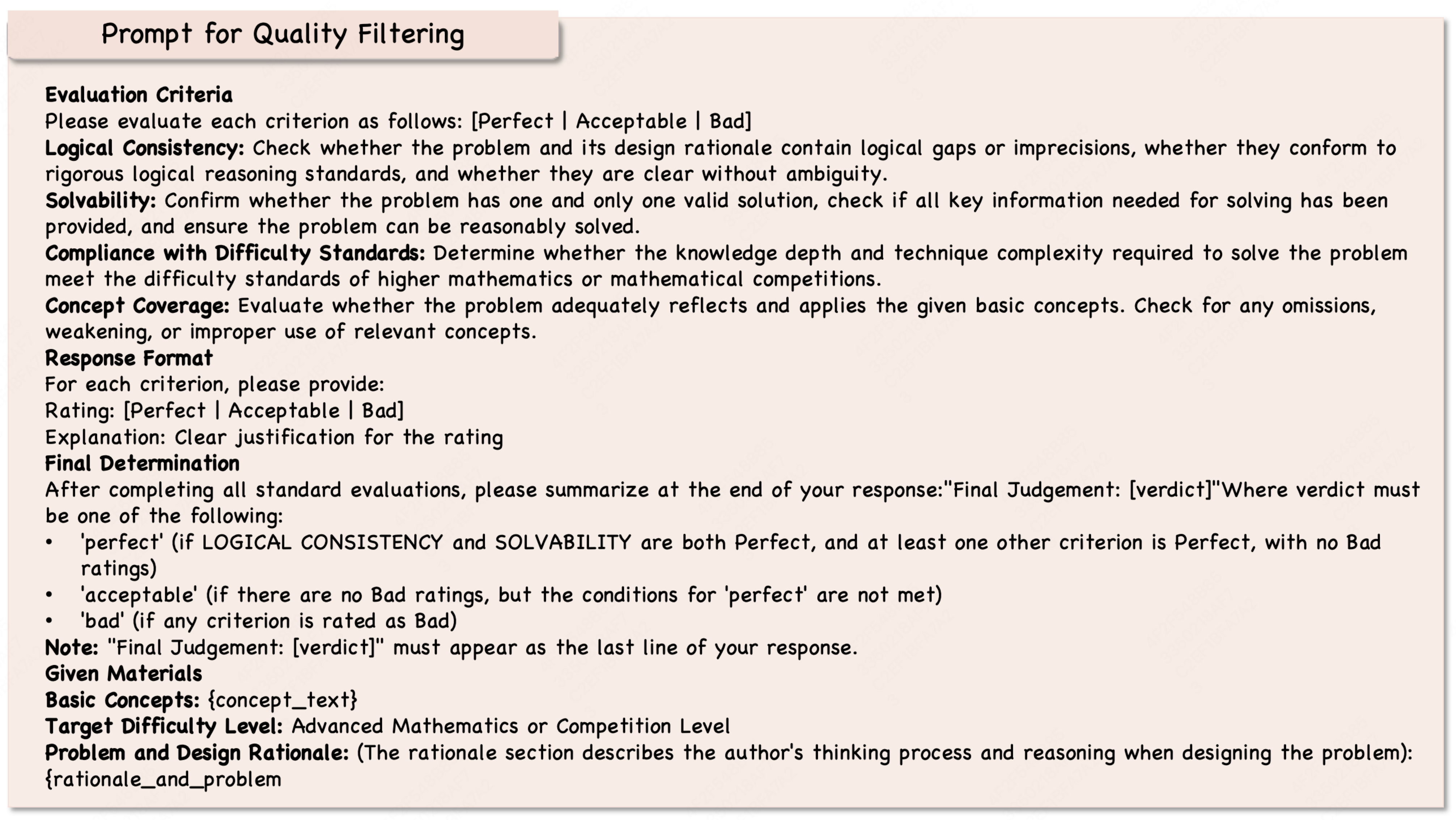}
    \caption{Prompt for quality filtering}
    \label{fig:quality-filter-prompt}
\end{figure*}

\begin{figure*}
    \includegraphics[width=1\linewidth]{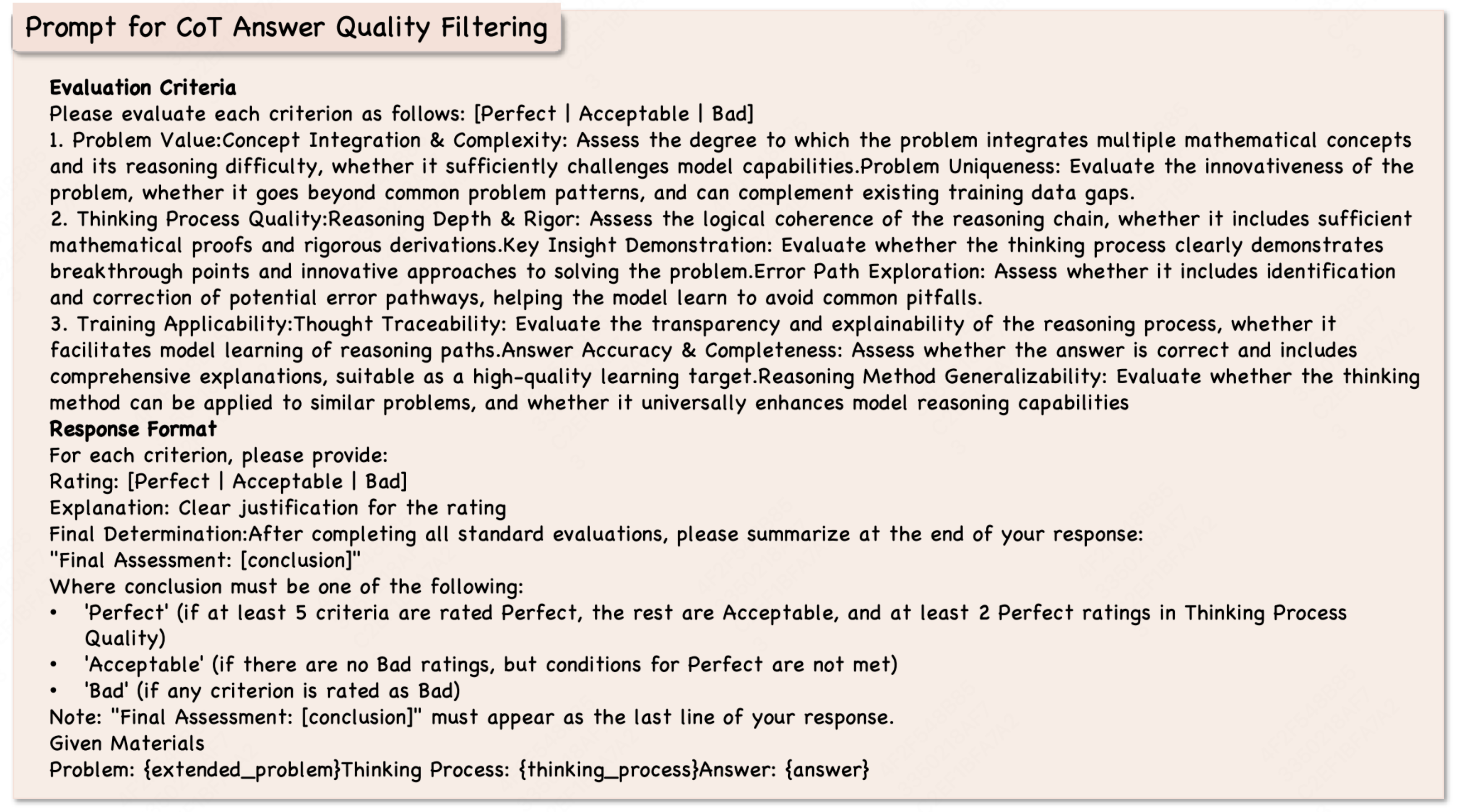}
    \caption{Prompt for CoT answer quality filtering}
    \label{fig:cot-quality-filter-prompt}
\end{figure*}

Figure \ref{fig:problem-generation-prompt} shows the prompt generating problems after we obtain the combinations of cognitive atoms. We apply strict filtering for both synthetic problems and generated answers, as shown in Figure \ref{fig:quality-filter-prompt} and \ref{fig:cot-quality-filter-prompt}.

\section{Case Study in Physics Generalization}
\label{app:physics_case_study_analysis}

\begin{figure*}[h!]
    \centering
    \includegraphics[width=\textwidth]{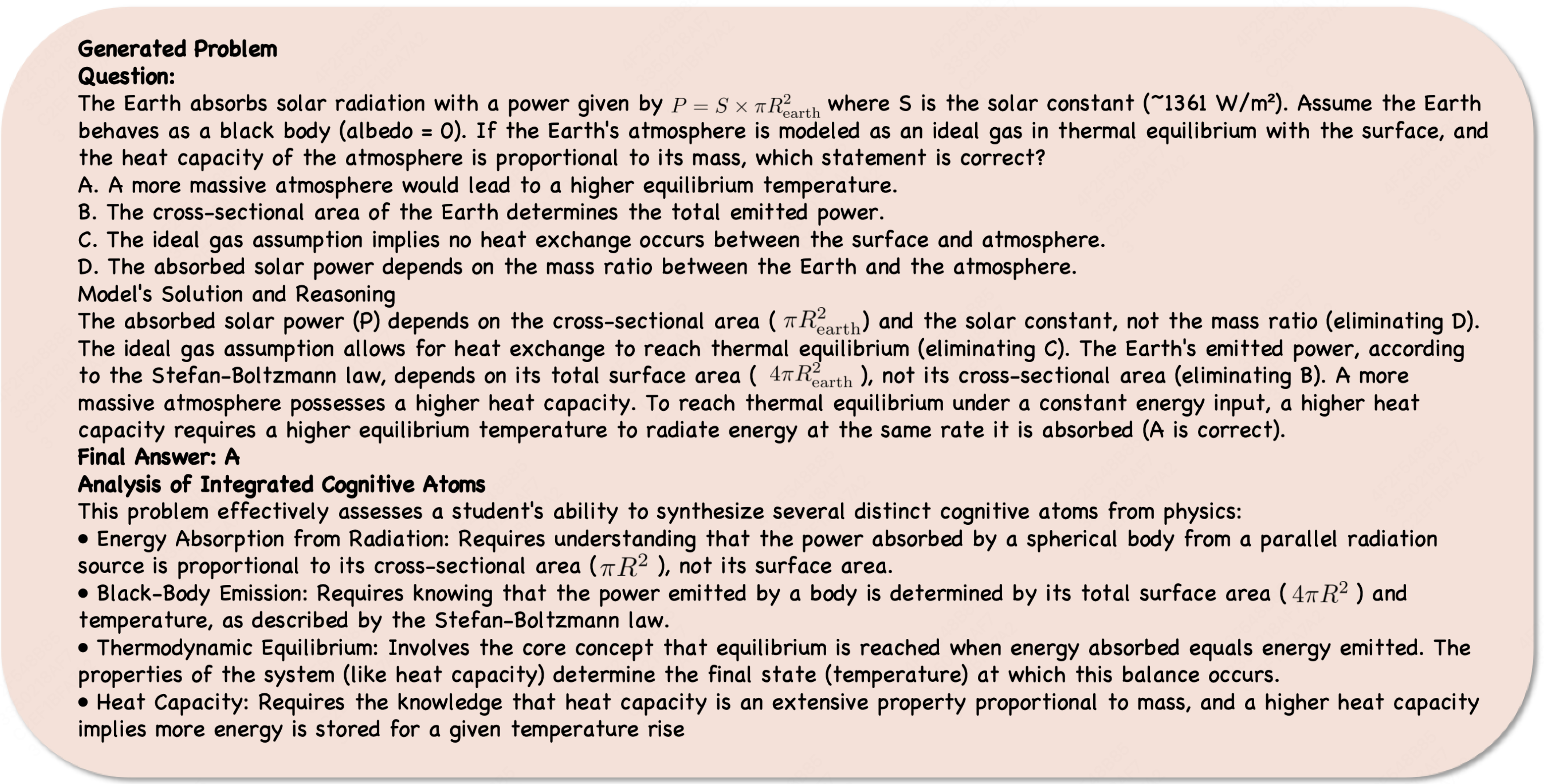}
    \caption{A case study illustrating the cross-domain generalization of CogAtom. The synthesized physics problem requires integrating distinct cognitive atoms from thermodynamics, black-body radiation, and mechanics.}
    \label{fig:physics_case_study}
\end{figure*}

Figure~\ref{fig:physics_case_study} presents a representative problem synthesized by CogAtom in the physics domain. This problem exemplifies a high-quality synthesis, as its solution requires the integration of four distinct cognitive atoms: Energy Absorption, Black-Body Emission, Thermodynamic Equilibrium, and Heat Capacity. The core reasoning challenge lies in reconciling the different geometric dependencies of energy absorption (proportional to the Earth's cross-sectional area) and thermal emission (proportional to the total surface area). A solver must correctly navigate this interplay and then deduce how a systemic property, such as heat capacity, influences the final equilibrium state. This demand for synthesizing a coherent model from competing principles is a hallmark of the cognitively complex problems our framework is designed to generate, built to stress-test and cultivate advanced reasoning capabilities.

\section{Case Study: Mathematical Problem Synthesis}
\label{app:math_case_study}

\begin{figure*}[t!]
\centering
\includegraphics[width=\textwidth]{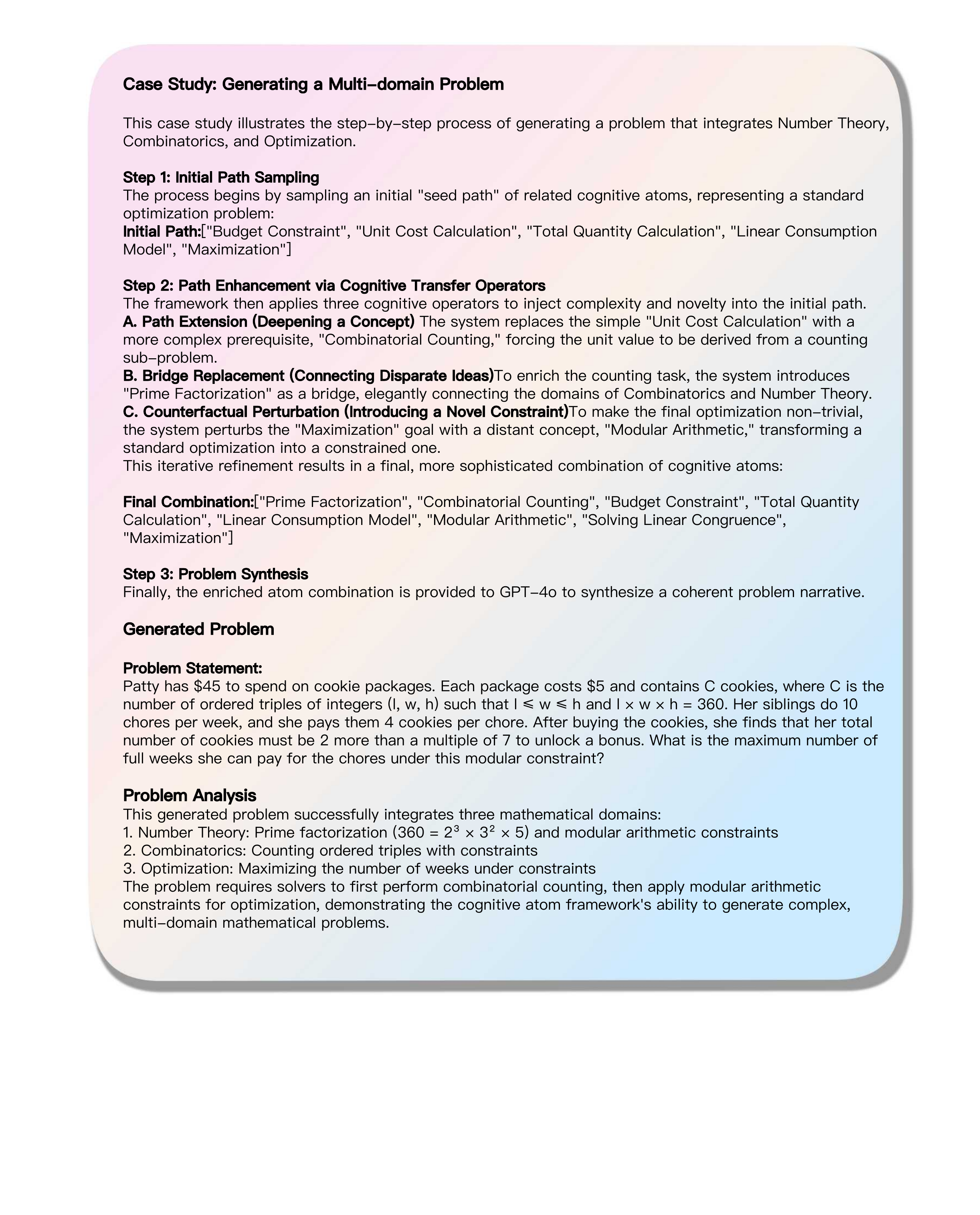}
\caption{A visualization of the Reasoning Chain Generation process for a multi-domain math problem. \textbf{(1) Path Sampling:} An initial, simple "seed path" focused on optimization is sampled from the global graph. \textbf{(2) Iterative Refinement:} The three Cognitive Transfer Operators—Path Extension, Bridge Replacement, and Counterfactual Perturbation—iteratively refine the path, injecting complexity by linking disparate domains like Number Theory and Combinatorics.}
\label{fig:case-study}
\end{figure*}

To provide a transparent, step-by-step illustration of our core Reasoning Chain Generation stage, this appendix presents a detailed case study. The process, visualized in Figure~\ref{fig:case-study}, deconstructs how our Cognitive Transfer Operators systematically transform a simple conceptual path into a sophisticated, multi-domain mathematical problem, showcasing the framework's capacity for controlled and creative synthesis.

\end{document}